\documentclass[conference]{IEEEtran}
\IEEEoverridecommandlockouts

\usepackage{cite}
\usepackage{amsmath,amssymb,amsfonts}
\usepackage{algorithmic}
\usepackage{graphicx}
\usepackage{textcomp}
\usepackage{xcolor}
\usepackage{booktabs}
\usepackage{algorithm}
\usepackage{algorithmic}
\usepackage{eso-pic}
\usepackage{graphicx}
\usepackage{tabularx}
\usepackage{subcaption}
\usepackage{hyperref}

\newcommand{\eat}[1]{}

\usepackage{xspace}

\def\BibTeX{{\rm B\kern-.05em{\sc i\kern-.025em b}\kern-.08em
    T\kern-.1667em\lower.7ex\hbox{E}\kern-.125emX}}

\hypersetup{
    pdftitle={An Efficient Model-Agnostic Approach for Uncertainty Estimation in Data-Restricted Pedometric Applications},
    pdfauthor={Viacheslav Barkov, Jonas Schmidinger, Robin Gebbers, Martin Atzmueller},
    pdfkeywords={machine learning, uncertainty estimation, pedometrics, digital soil mapping}
}

\begin{document}
\title{An Efficient Model-Agnostic Approach for Uncertainty Estimation in Data-Restricted Pedometric Applications\\
    \thanks{This work was supported by the Lower Saxony Ministry of Science and Culture (MWK), via the zukunft.niedersachsen program of the Volkswagen Foundation. Compute resources were funded by the Deutsche Forschungsgemeinschaft (DFG, German Research Foundation) project number 456666331.}}

\author{
    \IEEEauthorblockN{Viacheslav Barkov\IEEEauthorrefmark{1}\IEEEauthorrefmark{4},
        Jonas Schmidinger\IEEEauthorrefmark{1}\IEEEauthorrefmark{4},
        Robin Gebbers\IEEEauthorrefmark{4},
        Martin Atzmueller\IEEEauthorrefmark{1}\IEEEauthorrefmark{3}\IEEEauthorrefmark{7}}
    \IEEEauthorblockA{\IEEEauthorrefmark{1}Joint Lab Artificial Intelligence and Data Science, Osnabrück University, Osnabrück, Germany}
    \IEEEauthorblockA{Email: viacheslav.barkov@uni-osnabrueck.de, jonas.schmidinger@uni-osnabrueck.de}
    \IEEEauthorblockA{\IEEEauthorrefmark{4}Department of Agromechatronics, Leibniz Institute for Agricultural Engineering and Bioeconomy, Potsdam, Germany}
    \IEEEauthorblockA{Email: rgebbers@atb-potsdam.de}
    \IEEEauthorblockA{\IEEEauthorrefmark{3}Semantic Information Systems Group, Osnabrück University, Osnabrück, Germany}
    \IEEEauthorblockA{\IEEEauthorrefmark{7}German Research Center for Artificial Intelligence (DFKI), Osnabrück, Germany}
    \IEEEauthorblockA{Email: martin.atzmueller@uni-osnabrueck.de}
}

\maketitle

\begin{abstract}
    This paper introduces a model-agnostic approach designed to enhance uncertainty estimation in the predictive modeling of soil properties, a crucial factor for advancing pedometrics and the practice of digital soil mapping. For addressing the typical challenge of data scarcity in soil studies, we present an improved technique for uncertainty estimation. This method is based on the transformation of regression tasks into classification problems, which not only allows for the production of reliable uncertainty estimates but also enables the application of established machine learning algorithms with competitive performance that have not yet been utilized in pedometrics.

    Empirical results from datasets collected from two German agricultural fields showcase the practical application of the proposed methodology. Our results and findings suggest that the proposed approach has the potential to provide better uncertainty estimation than the models commonly used in pedometrics.
\end{abstract}


\section{Introduction}
Soils are fundamental to agriculture and ecosystems, supporting plant growth, water filtration, and carbon sequestering~\cite{keesstra2016forum}. Understanding the complexity of soils through statistical inferences is the essence of pedometrics. One main goal of pedometrics is to create reliable digital soil maps which can guide ecological and agricultural decisions~\cite{mcbratney1986introduction}. Digital soil mapping utilizes predictive learning algorithms, drawing upon data from proximal and remote sensing to map soil characteristics, a process which is essential for advancing sustainable agricultural practices and environmental health~\cite{mcbratney2003digital}.

In recent years, machine learning has emerged as a powerful tool in pedometrics, standing out for its strong predictive power~\cite{wadoux2020machine}. However, the accurate prediction of soil properties is often challenged by a lack of extensive data due to the expensive and labor-intensive process of soil sampling~\cite{pennock2007soil}. This often results in smaller datasets with too few training samples, limiting the performance of predictive models.

In addition to data scarcity, uncertainty in model predictions is a critical factor in digital soil mapping, especially for users like farmers who rely on these predictions for decision-making~\cite{heuvelink2018uncertainty}. Reliable uncertainty measures are necessary to build confidence in model outputs and support informed actions. Therefore, it is essential to quantify the respective prediction uncertainty to properly demonstrate the capabilities of a predictive model.

To address these challenges, this paper introduces a model-agnostic approach for uncertainty estimation in pedometrics.
The proposed approach enables models to directly output uncertainty estimates, eliminating the need for a separate calibration dataset required by established model-agnostic uncertainty estimation methods in pedometrics. This prevents the further reduction of the training dataset size, which may prove advantageous in scenarios of data scarcity. Additionally, this approach broadens the toolkit of machine learning models available for pedometrics, introducing established algorithms that have not yet been utilized in this field.

The proposed method is validated on predicting soil organic carbon (SOC) and pH with two distinct field-scale datasets collected in Germany, highlighting its effectiveness compared to traditional methods. The results indicate the efficacy of this approach to offer improved uncertainty estimation in pedometrics, contributing to the progressive development of digital soil mapping.

\section{Background}

\begin{figure*}[htbp]
    \centering
    \begin{subfigure}{.5\textwidth}
        \centering
        \includegraphics[width=\linewidth]{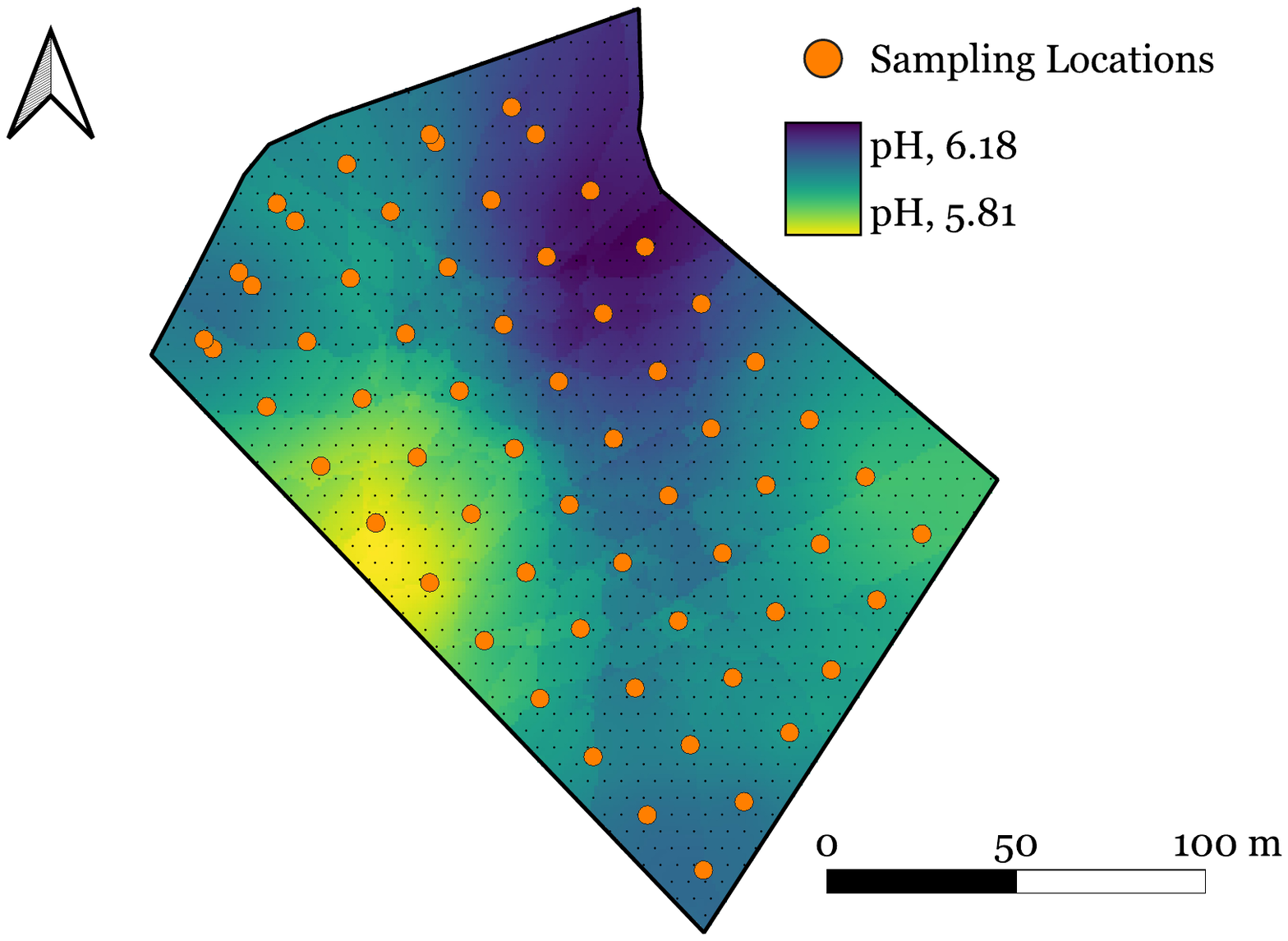}
        \caption{Predicted pH values}
        \label{fig:dsm_ph}
    \end{subfigure}%
    \begin{subfigure}{.5\textwidth}
        \centering
        \includegraphics[width=\linewidth]{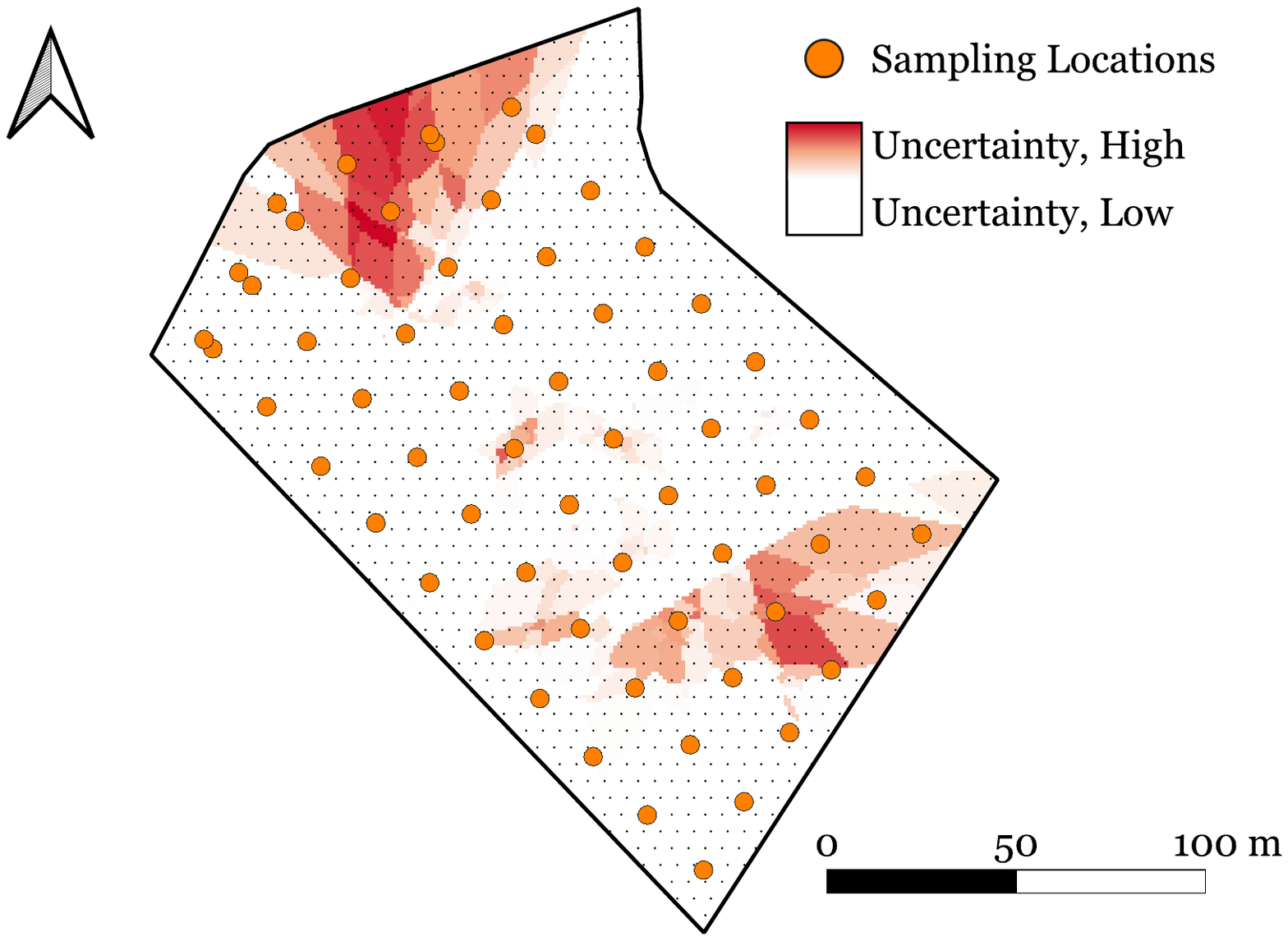}
        \caption{Model uncertainty estimates}
        \label{fig:dsm_uncertainty}
    \end{subfigure}
    \caption{Digital soil map showing predicted soil pH values using TabPFN and model uncertainty estimates with the proposed method for the Boelingen dataset, with ordinary kriging applied to interpolate point predictions into rasters.}
    \label{fig:dsm}
\end{figure*}

\subsection{From Regression to Classification Task}
Converting regression problems into classification tasks is a well-established method in structured tabular data analysis~\cite{torgo1996regression,weiss1995rule}. Despite the intrinsic continuous nature of certain problems, a transformation into a classification framework frequently leads to enhanced model performance, keeping this method a subject of ongoing research~\cite{zhang2023improving,stewart2023regression,pintea2023step,farebrother2024stop}.
This method generally involves dividing the continuous output range into discrete intervals, a technique often referred to in literature as binning. Each bin is treated as a distinct class, and the classification model is trained to minimize the categorical cross-entropy instead of the traditional mean-squared error typically used in regression problems~\cite{farebrother2024stop}.

The approach of framing regression problems as classification tasks has proven valuable not only in tabular data tasks but also in a diverse range of different fields within the realm of machine-learning research, such as computer vision and reinforcement learning.
In the field of computer vision, the benefit of applying this approach has been practically demonstrated with diverse set of tasks such as age estimation from a single face image~\cite{rothe2018deep}, human pose estimation~\cite{rogez2019lcr}, automatic colorization of grayscale photographs~\cite{zhang2016colorful}.
The domain of reinforcement learning also showcases the versatility of this approach by applying it to tasks such as robotic manipulation~\cite{farebrother2024stop,akkaya2019solving}, Atari games and board games agents\cite{farebrother2024stop,schrittwieser2020mastering}, language-agent tasks\cite{farebrother2024stop}.

Besides the potential increase in the model predictive performance, multiple other advantages of adopting a classification approach over regression have been identified in the literature, such as improved learning of feature representations~\cite{zhang2023improving}, implicit bias in gradient-based optimization methods when employing classification losses such as cross-entropy~\cite{stewart2023regression}, dealing with imbalanced datasets~\cite{pintea2023step}, improved scalability~\cite{farebrother2024stop}, or enhanced robustness of models against noisy or non-stationary data~\cite{farebrother2024stop}.

One of other principal advantages of employing the classification approach is the possibility of applying a wide range of established off-the-shelf machine learning algorithms, especially including those not typically available for regression tasks. For instance, one of the earliest successful applications of classical algorithms for classification to achieve enhanced predictive performance utilized models such as the C4.5 decision tree model~\cite{quinlan1996bagging} and the CN2 rule induction algorithm~\cite{clark1987induction} with various datasets from the UCI machine learning repository~\cite{torgo1996regression}, demonstrating their effectiveness. In this context, the introduction of TabPFN~\cite{hollmann2022tabpfn} into regression tasks represents one of the important contributions of this paper. Designed for small tabular classification problems, TabPFN is a transformer-based model which achieves competitive results~\cite{mcelfresh2024neural} almost instantly without necessitating extensive parameter tuning. Importantly, TabPFN is a variant of Prior-Data Fitted Networks (PFNs)~\cite{muller2021transformers}. PFNs have the potential to offer considerable benefits in data-scarce situations, which makes their application highly relevant in the field of pedometrics where this is a common issue. PFNs are essentially deep learning models designed to approximate the posterior predictive distribution through a process of pre-training on synthetic data generated from a specific prior, which can help to avoid the need for additional model training when only small real-world datasets are available. This pre-training could embed the prior knowledge into the model, enabling it to make informed predictions even when only limited data is available for direct inference. The successful application of TabPFN is particularly noteworthy given that traditional machine learning often proves more effective than neural networks on tabular datasets with limited training data ~\cite{grinsztajn2022tree}.

Additionally, an important and yet under-investigated advantage of the classification-based approach to regression tasks is due to its inherent capability to establish a measure of uncertainty~\cite{li2021deep}. By discretizing the prediction space into bins and estimating the probabilities for each separate bin, this approach allows to capture the variation and uncertainty of model predictions. This is a critical improvement over traditional regression methods, which typically only offer point estimates. An important contribution of this paper is a further development of this approach providing an improved methodology that integrates uncertainty estimation into predictive modeling for real-world application in pedometrics, thus improving robustness and  interpretability of the model predictions.

\subsection{Predictive Uncertainty Estimation}

Uncertainty in the context of machine learning refers to the inherent limitations in predictive models due to the imperfections within the model itself as well as the limitations within the input data~\cite{gawlikowski2023survey}. These uncertainties are generally classified as either data uncertainty, also referred as statistical or aleatoric uncertainty, or model uncertainty, also referred as systemic or epistemic uncertainty~\cite{hullermeier2021aleatoric}. Epistemic uncertainty can arise from a variety of factors, including errors in the model's training process, structural insufficiencies of the model architecture, or unfamiliarity with certain samples due to poor representation in the training dataset. In contrast, aleatoric uncertainty is inherent to the data itself and results from noise or variance in the input data that cannot be diminished or learned away by the model~\cite{hullermeier2021aleatoric}. Model uncertainty estimation refers to the process of quantifying the uncertainty that is specific to the model in question~\cite{gawlikowski2023survey}. The ability to estimate uncertainty in model predictions is crucial in real-world applications, as point predictions, while being informative, do not offer insights into the associated confidence or reliability of these predictions~\cite{schmidinger2023validation,kasraei2021quantile}.

The ability to estimate model uncertainty provides distinct advantages to end users, particularly within the context of pedometrics and digital soil mapping. Estimating uncertainty enhances the practical application and interpretability of soil maps. This allows for informed decisions in agriculture and land management~\cite{heuvelink2018uncertainty}. For instance, providing prediction intervals alongside the predicted soil map aids users in understanding the reliability and accuracy of the soil map, facilitating decisions on whether the predictions are adequate for their intended use~\cite{schmidinger2023validation}. Moreover, uncertainty prediction enhances the integration of soil maps into decision-making processes by visually indicating the accuracy of predictions on the map and highlighting potential risks in their application~\cite{schmidinger2023validation}. An example of the visual representation of uncertainty estimates in digital soil mapping is presented in Fig.~\ref{fig:dsm}.

Another important yet often overlooked benefit of uncertainty estimation is its ability to detect outliers, deviations, and data drift within the gathered data. When predictive models are developed for soil properties, it is crucial to take into account the heterogeneous nature of soils across different fields and over time~\cite{lagacherie2020analysing}. Data drift detection is integral for maintaining the accuracy of soil property predictions, ensuring that the predictive models remain accurate and reflect current conditions. Uncertainty estimation is a valuable tool for drift detection because it directly measures the confidence of model predictions by leveraging the model's inherent margin or region of uncertainty~\cite{sethi2017reliable}. This makes it an effective indicator of when a model may start to fail due to changes in the underlying data distribution~\cite{sethi2017reliable}. Early detection of data drift can help identify when digital soil mapping model assumptions no longer hold true, prompting timely interventions to investigate these anomalies. This can improve the reliability of soil maps for the end user, ensuring that predictive models for soil properties remain accurate and represent current conditions.

The importance of uncertainty estimation in the context of soil mapping has been recognized in the literature, with several studies proposing model-agnostic methods for quantifying uncertainty in soil property predictions~\cite{kasraei2021quantile,kakhani2024uncertainty}. Among these, Quantile Regression post-processing has been demonstrated as a generic approach applicable across a variety of predictive models and different soil properties~\cite{kasraei2021quantile}. Conformal Prediction is another generic approach recognized for its capacity to output statistically valid prediction intervals without making heavy assumptions about the data~\cite{kakhani2024uncertainty}. However, these methods have their drawbacks, particularly when dealing with the commonplace issue of data scarcity in soil studies. The reliance of both these approaches on further splitting training datasets into separate calibration and training sets leads to an even smaller quantity of data available for model training, which may impair the model's performance and the quality of its uncertainty estimates. Although there exist techniques like Quantile Regression Forests~\cite{vaysse2017using} that bypass the need for calibration datasets, they are tied to a specific model and thereby do not retain the advantage of model-agnostic methods. This can be a significant limitation, highlighting the need for flexible, model-agnostic approaches that enable the use of existing off-the-shelf machine learning algorithms without relying on a separate calibration dataset. Addressing this need is one of the important contributions of this paper, which presents an uncertainty estimation approach that overcomes these limitations.

\section{Methodology}
\subsection{Method Overview}
We introduce a model-agnostic approach for uncertainty estimation in regression tasks and demonstrate its utility in the context of pedometrics and digital soil mapping. To achieve this, we develop a universal adapter designed to seamlessly transform regression tasks into classification problems, enabling the use of classification algorithms for regression. First, the universal adapter discretizes the continuous target variable from the training dataset into distinct intervals, identified as bins. Following the discretization, the classification model is trained. Subsequently, the adapter reconstructs the continuous prediction by leveraging the probabilistic output of the trained classifier. Additionally, with the aim to quantify the model uncertainty, the adapter calculates the standard deviation of the bin structure. This standard deviation serves as a proxy for the model's intrinsic uncertainty, reflecting the variability within the bin predictions. The challenge of selecting the appropriate number of bins and the binning strategy is addressed by employing an ensemble approach that combines predictions from models trained with various bin sizes and binning strategies. We refer to this combined approach that integrates all the separate steps as Binned Uncertainty Estimation Ensemble.

\subsection{Target Discretization}
To convert the continuous regression problem into a discrete classification task, we discretize the continuous output variable $y$ into a finite set of intervals $\{B_1, B_2, \ldots, B_K\}$ called bins. For this, we employ different binning strategies.

Different binning strategies dictate how to group the continuous values into categorical bins. We employ two common binning strategies: Uniform Binning and Quantile Binning.

\paragraph{Uniform Binning}
Under Uniform Binning, the continuous range of $y$ is divided into $k$ intervals of equal length. This results in bin boundaries being defined as equidistant points between the lowest and highest observed values of $y$. Let \( y_{\min} \) and \( y_{\max} \) denote the minimum and maximum values of the data, respectively. Then each bin has a width of \( \Delta = \frac{y_{max} - y_{min}}{k} \). The bin edges are given by:
\begin{equation}
    b_{i} = y_{\min} + i \cdot \Delta \quad \text{for} \quad i = 0, 1, 2, \ldots, k
\end{equation}
Each data point \( y \) is assigned to the bin \( j \) such that:
\begin{equation}
    b_{j - 1} < y \leq b_{j} \quad
\end{equation}

\paragraph{Quantile Binning}
Quantile Binning partitions the data into $k$ bins, each containing approximately the same number of data points. For a given empirical data distribution of $y$, bins are positioned at quantiles of the distribution, which ensures evenly populated bins.
If \( Q(p) \) indicates the \( p \)-quantile of the dataset, the bin edges are defined as:
\begin{equation}
    b_{i} = Q\left(\frac{i}{k}\right) \quad \text{for} \quad i = 0, 1, 2, \ldots, k
\end{equation}
Each data point \( y \) is then assigned to the bin \( j \) such that:
\begin{equation}
    b_{j - 1} < y \leq b_{j} \quad
\end{equation}

\subsection{Classification Predictions to Continuous Estimates}
To facilitate the transition from classification to regression, the predicted class probabilities are converted back into a regression estimate $\hat{y}$. Given the predicted probabilities $\mathbf{p}$ over the bins, the predictive mean $\hat{y}$ is estimated as a weighted sum of the bin midpoints $\{m_1, m_2, \ldots, m_K\}$:

\begin{equation}
    \hat{y}(\mathbf{x}) = \sum_{k=1}^K p_k \cdot m_k,
\end{equation}

where $p_k = P(B_k|\mathbf{x})$ is the predicted probability of bin $B_k$ for input feature vector $\mathbf{x}$, and $m_k$ is the midpoint of bin $B_k$.

\subsection{Estimation of Model Uncertainty}
To estimate the intrinsic model uncertainty, we calculate the standard deviation $\sigma(\mathbf{x})$ of the predicted probabilities across the bins. For a new input $\mathbf{x} \in \mathbb{R}^D$, the model uncertainty is given by:

\begin{equation}
    \sigma(\mathbf{x}) = \sqrt{\sum_{k=1}^K p_k \cdot (m_k - \hat{y}(\mathbf{x}))^2},
\end{equation}

where $\hat{y}(\mathbf{x})$ is the weighted sum of bin midpoints representing the predicted regression value for the adapter, and $p_k$ represents the probability that the true value of $y$ falls within bin $B_k$ centered at $m_k$.

\subsection{Addressing Bin Size Selection by Ensemble Modeling}
Selecting the optimal bin size $K$ and binning strategy is a challenging task that could significantly impact the performance of a model. One potential approach is to integrate varying bin sizes and strategies within a hyperparameter grid search. However, expanding the search space with these variables can lead to a substantial increase in computational complexity due to the exponential growth in the number of model configurations. To address these challenges, we employ an ensemble model that consolidates predictions from a series of base models. Given a new sample $\mathbf{x} \in \mathbb{R}^D$, the ensemble prediction $\hat{y}_{\text{ensemble}}(\mathbf{x})$ is calculated as the weighted sum over a range of bin configurations:

\begin{equation}
    \hat{y}_{\text{ensemble}}(\mathbf{x}) = \sum_{b=1}^B w_b \sum_{k=1}^{K_b} p_k^b(\mathbf{x}) \cdot m_k^b,
\end{equation}

where $B$ is the total number of bin configurations considered, $K_b$ represents the number of bins in the $b$-th configuration, $w_b$ is the associated weight of the $b$-th base model, $p_k^b(\mathbf{x})$ is the predicted probability that $\mathbf{x}$ belongs to the $k$-th bin as estimated by the $b$-th model, and $m_k^b$ indicates the midpoint value of the $k$-th bin in the $b$-th model.

By incorporating a diverse selection of bin sizes and strategies within the ensemble, the proposed ensemble model reduces the dependency on the appropriate bin size selection. This not only simplifies the model training process but also has the potential to yield more robust point predictions and uncertainty estimates.

\begin{table}[tbp]
    \renewcommand{\arraystretch}{1.5}
    \centering
    \caption{Hyperparameter Grid for Utilized Algorithms}
    \label{table_grid_search}
    \begin{tabular}{p{0.3\linewidth} p{0.6\linewidth}}
        \toprule
        \textbf{Algorithm} & \textbf{Parameter Search Grid}                   \\
        \midrule
        CatBoost           &
        learning\_rate: \{0.005, 0.01, 0.05, 0.1\}\newline
        depth: \{4, 6, 8, 10\}\newline
        subsample: \{0.6, 0.8, 1.0\}\newline
        l2\_leaf\_reg: \{3.0, 10.0\}                                          \\
        \addlinespace
        XGBoost            &
        learning\_rate: \{0.005, 0.01, 0.05, 0.1\}\newline
        max\_depth: \{4, 6, 8, 10\}\newline
        subsample: \{0.6, 0.8, 1.0\}\newline
        colsample\_bytree: \{0.5, 0.75, 1.0\}\newline
        gamma: \{0.5, 1, 2\}                                                  \\
        \addlinespace
        RandomForest       &
        max\_depth: \{4, 6, 8, 10\}\newline
        max\_features: \{1.0, 'sqrt', 'log2'\}\newline
        min\_samples\_leaf: \{1, 2, 4\}\newline
        max\_samples: \{0.6, 0.8, 1.0\}                                       \\
        \addlinespace
        QuantileRegression & alpha: \{0.1, 0.5, 1.0, 2.0, 5.0, 10.0\}\newline
        fit\_intercept: \{True, False\}                                       \\
        \bottomrule
    \end{tabular}
\end{table}

\section{Experiments}
\subsection{Data}
SOC and pH values were measured in soil samples from two distinct agricultural fields in Germany. The first study site, referred to as the GW field, is located in north-east Germany, with an area of about 3.4 hectares~\cite{vogel2022direct}. The second site of interest, refered to as the Boelingen field, is located in south-west Germany, encompassing an area of approximately 2.7 hectares~\cite{tavakoli2022rapidmapper}. Soil samples are gathered through regular grid sampling, with 144 samples for the GW field and 62 samples for the Boelingen field. Predictors are retrieved from various sensors, including a multispectral bare soil image (Sentinel-2) and terrain parameters from remote sensing as well as gamma-ray, electrical conductivity, and pH-proxies from in-situ proximal soil sensing platforms.

To spatially align sensor data with the corresponding soil properties, ordinary kriging was employed. Ordinary kriging is one of the primary geostatistical methods used as a spatial interpolation technique in digital soil mapping~\cite{hengl2004generic}. This approach utilizes the spatial autocorrelation structure of variables to determine interpolation weights and estimate values at unsampled locations using weighted averages of nearby sampled points. Ordinary kriging enables the establishment of point-by-point relationships between laboratory-analyzed soil samples and the nearest sensor readings, addressing challenges such as irregular sensor locations and the integration of data from multiple sensors with different resolutions prior to implementing main modeling approaches.

\subsection{Experimental Setup}
To evaluate the performance of our proposed method, we compare it against established state-of-the-art techniques commonly applied in pedometrics, specifically Quantile Regression post-processing and Conformal Prediction. Within this framework, we apply the adapted TabPFN model along with three popular machine learning algorithms, namely CatBoost, XGBoost, and RandomForest. Furthermore, we assess the capability of our proposed method in relation to regular Quantile Regression as a one-stage method, which is implemented without  additional methods for uncertainty estimation.

\begin{algorithm}[tbp]
    \caption{Algorithm of K$\times$N nested cross-validation with extensive hyperparameter search.}
    \label{alg:nested_cv}
    \begin{algorithmic}
        \REQUIRE Dataset $D$, number of outer folds $K$, number of inner folds $N$, learning algorithm $A$, hyperparameter grid $G$, need for calibration set $C$
        \FOR{$i=1$ \TO $K$}
        \STATE Split $D$ into $D_{\text{train}}^i$ and $D_{\text{test}}^i$
        \STATE Initialize ensemble $E_i$ to an empty set
        \FOR{$j=1$ \TO $N$}
        \STATE Split $D_{\text{train}}^i$ into $D_{\text{train}}^{i,j}$ and $D_{\text{val}}^{i,j}$
        \IF{$C$}
        \STATE Further split $D_{\text{train}}^{i,j}$ to create a calibration set $D_{\text{cal}}^{i,j}$
        \ELSE
        \STATE Proceed without calibration set
        \ENDIF
        \FORALL{hyperparameter combinations $\theta$ in grid $G$}
        \STATE Train algorithm $A$ on $D_{\text{train}}^{i,j}$ with $\theta$
        \IF{$C$}
        \STATE Calibrate $A$ on $D_{\text{cal}}^{i,j}$
        \ENDIF
        \STATE Validate $A$ on $D_{\text{val}}^{i,j}$
        \STATE Record validation performance
        \ENDFOR
        \STATE Select best hyperparameter set $\theta_{\text{best}}^{i,j}$
        \STATE Train new model $M_{ij}$ with $D_{\text{train}}^{i,j}$ and $\theta_{\text{best}}^{i,j}$
        \IF{$C$}
        \STATE Calibrate model $M_{ij}$ on $D_{\text{cal}}^{i,j}$
        \ENDIF
        \STATE Add $M_{ij}$ to ensemble $E_i$
        \ENDFOR
        \STATE Use ensemble $E_i$ to predict $D_{\text{test}}^i$
        \ENDFOR
    \end{algorithmic}
\end{algorithm}

The hyperparameters for each algorithm are optimized using an exhaustive grid search strategy, with the best-performing set selected based on the validation results. The hyperparameter grids for each algorithm are detailed in Table~\ref{table_grid_search}.

For each model, we conduct a nested cross-validation procedure with $K=5$ outer folds and $N=5$ inner folds. The outer folds are used to evaluate the model's generalization performance, while the inner folds are employed for hyperparameter tuning. The performance of each model is assessed using the Continuous Ranked Probability Score (CRPS), a well-established metric for assessing probabilistic forecasts.

For the methods that require a calibration set, we split the training data into a training subset and a calibration subset in each inner fold. The calibration set is not used for training the model and is employed to calibrate the uncertainty estimation models like Quantile Regression post-processing and Conformal Prediction. In cases where the model does not require a calibration set, like the proposed Binned Uncertainty Estimation Ensemble and one-stage Quantile Regression approach, the training data is used directly for model training.

For the outer fold predictions, the ensemble approach is employed by aggregating predictions from each inner fold model. This way the predictions of unseen outer fold test sets are obtained from a diverse ensemble of inner fold models with optimal parameters determined by grid search on inner fold validation set.
The algorithm of the nested cross-validation procedure is shown in Algorithm~\ref{alg:nested_cv}.

\subsection{Evaluation Metric}
To assess the performance of the proposed method, we employ the CRPS, a widely recognized metric for evaluating probabilistic predictions. CRPS has seen application in various fields, including meteorology, pedometrics and digital soil mapping, reflecting its versatility and effectiveness as a performance metric for probabilistic models~\cite{schmidinger2023validation,gneiting2007strictly}.

\begin{table*}[tbp]
    \renewcommand{\arraystretch}{1.5}
    \caption{Comparison of Models for pH and SOC on Boelingen and GW Datasets Using CRPS}
    \label{tab:table_results}
    \centering
    \begin{tabular}{@{}ccccccc@{}}
        \toprule
        Model               & Uncertainty Estimation Method                              & \multicolumn{2}{c}{Boelingen} & \multicolumn{2}{c}{GW}                                     \\
        \cmidrule(lr){3-4} \cmidrule(l){5-6}
                            &                                                            & pH, CRPS                      & SOC, CRPS              & pH, CRPS        & SOC, CRPS       \\
        \midrule
        TabPFN              & \textit{Binned Uncertainty Estimation Ensemble (Proposed)} & 0.1096                        & \textbf{0.1299}        & 0.2162          & \textbf{0.0843} \\
        XGBoost             & \textit{Binned Uncertainty Estimation Ensemble (Proposed)} & 0.1089                        & 0.1344                 & 0.2153          & 0.0881          \\
        CatBoost            & \textit{Binned Uncertainty Estimation Ensemble (Proposed)} & \textbf{0.1029}               & 0.1351                 & \textbf{0.2131} & 0.0846          \\
        RandomForest        & \textit{Binned Uncertainty Estimation Ensemble (Proposed)} & 0.1060                        & 0.1341                 & 0.2221          & 0.0859          \\
        XGBoost             & Quantile Regression Post-Processing                        & 0.1350                        & 0.1657                 & 0.2208          & 0.0933          \\
        CatBoost            & Quantile Regression Post-Processing                        & 0.1235                        & 0.1456                 & 0.2214          & 0.0873          \\
        RandomForest        & Quantile Regression Post-Processing                        & 0.1180                        & 0.1638                 & 0.2186          & 0.0884          \\
        XGBoost             & Conformal Prediction                                       & 0.1288                        & 0.1432                 & 0.2241          & 0.0975          \\
        CatBoost            & Conformal Prediction                                       & 0.1112                        & 0.1334                 & 0.2261          & 0.0853          \\
        RandomForest        & Conformal Prediction                                       & 0.1130                        & 0.1435                 & 0.2252          & 0.0853          \\
        Quantile Regression & --                                                         & 0.1229                        & 0.1492                 & 0.2613          & 0.0879          \\
        \bottomrule
    \end{tabular}
\end{table*}

CRPS is defined as the integral of the squared difference between the predictive cumulative distribution functions (CDFs) and empirical CDFs generated from single test data samples. It is comparable to traditional point prediction metrics like the mean squared error, but evaluates the entire predictive CDF. This metric assesses the predictive ability of probabilistic models by evaluating the sharpness, or the distribution width, of the predictive CDFs, as well as their reliability~\cite{gneiting2007strictly}. CRPS provides a robust measure for validating uncertainty predictions by analyzing the whole probability distribution, thus eliminating the shortcomings of metrics like PICP that fail to identify issues with distribution sharpness or one-sided biases~\cite{schmidinger2023validation}.

Additionally, a strong advantage of the CRPS is that it includes predictive performance of the model as part of its evaluation metric. CRPS is sensitive to the accuracy of the predicted distributions as well as the distance between forecasted outcomes and actual observations, which enables the assessment of both model performance and uncertainty estimation performance with a single consistent metric~\cite{gneiting2007strictly}.

\subsection{Results}
Table~\ref{tab:table_results} presents a summary of the model performance for predicting pH and SOC on the Boelingen and GW datasets, evaluated using CRPS. The results are a comparison of TabPFN, XGBoost, CatBoost, Random Forest and Quantile Regression models with different uncertainty estimation techniques, including the proposed Binned Uncertainty Estimation Ensemble, Quantile Regression post-processing, and Conformal Prediction.
The proposed Binned Uncertainty Estimation Ensemble method provides competitive CRPS across both datasets for pH and SOC indicators.

For the Boelingen dataset, the CRPS values for pH and SOC range from 0.1029 to 0.1350 and 0.1299 to 0.1657, respectively, while the CRPS values for pH and SOC models paired with Binned Uncertainty Estimation Ensemble method range from 0.1029 to 0.1096 and 0.1299 to 0.1351, respectively. For the GW dataset, the CRPS values for pH and SOC range from 0.2131 to 0.2613 and 0.0843 to 0.0975, respectively, while the CRPS values for pH and SOC models paired with Binned Uncertainty Estimation Ensemble method range from 0.2131 to 0.2221 and 0.0843 to 0.0881, respectively. The results indicate that the TabPFN model, when paired with Binned Uncertainty Estimation Ensemble, consistently outperforms the other models across both datasets for SOC predictions uncertainty. The CatBoost model paired with Binned Uncertainty Estimation Ensemble demonstrates the best performance for pH predictions uncertainty across both datasets.

Overall, the experiments' findings indicate that the proposed Binned Uncertainty Estimation Ensemble method, particularly when paired with TabPFN and CatBoost, is effective for estimating model uncertainty in the prediction of soil attributes such as pH and SOC, with consistent performance as indicated by achieving the lowest CRPS values across all experiments.

A visual representation of the proposed uncertainty estimation method applied to the Boelingen dataset is showcased in Fig.~\ref{fig:dsm}. Specifically, Fig.~\ref{fig:dsm_ph} displays the spatial distribution of pH values as predicted by the TabPFN model adapted for the regression task using the proposed binning approach. Fig.~\ref{fig:dsm_uncertainty} illustrates the model's uncertainty estimates, highlighting areas with varying levels of prediction confidence. The combination of these maps serves two important purposes. First, it enables the identification of potential management zones within the field based on pH distribution. Second, it pinpoints regions that may require additional soil sampling or more focused management due to higher model uncertainty in pH predictions. This visual representation of both the target variable distribution and the associated uncertainty estimates is particularly valuable in practical settings such as precision agriculture, where informed spatial decision-making is crucial for optimizing field management strategies~\cite{breure2022loss}.

\section{Conclusion}

This paper introduced an efficient model-agnostic approach for uncertainty estimation and showcased its potential utility in the realm of pedometrics and digital soil mapping. The proposed method transforms regression tasks into classification problems, allowing the application of established machine learning algorithms with competitive performance that have not yet been utilized in pedometrics. The presented approach offers an effective technique for uncertainty estimation in data-restricted applications, addressing the challenges of data scarcity and providing reliable uncertainty estimates.

The potential for applying novel models through the proposed transformation from regression to classification tasks has been demonstrated using TabPFN, a transformer-based model with competitive results for classification tasks in various domains with data scarcity issues. The adoption of this model is particularly noteworthy, as traditional machine learning methods often outperform neural networks on tabular datasets with limited sample sizes~\cite{mcelfresh2024neural}. TabPFN stands out due to its pre-training, which involves approximating Bayesian inference on synthetic datasets, allowing the model to be successfully applied in scenarios with limited data. Another considerable advantage of TabPFN is the absence of a need for a hyperparameter search, an important step when applying boosting algorithms and decision trees. This feature makes the application of TabPFN faster and easier than that of other machine learning algorithms.

The potential in uncertainty estimation was demonstrated by comparing the proposed method with traditional model-agnostic approaches currently used in digital soil mapping and pedometrics, like Quantile Regression post-processing and Conformal Prediction. The results indicate that the proposed method might have the potential to provide better uncertainty estimation in situations with scarce training data. The results were validated on two distinct datasets collected in east and west of Germany. The method achieved lower CRPS scores on both datasets for both pH and SOC, showing the best results when combined with the TabPFN and CatBoost models.

Our empirical results demonstrate the proposed approach as being versatile and efficient for uncertainty estimation in pedometrics, enabling the use of a diverse range of machine learning algorithms and enhancing the reliability of model predictions. In general, accurate estimates of uncertainty are essential for informed decision-making in fields like agriculture and land management, where they can facilitate better risk assessment and resource management.

Future work will focus on extending the proposed approach, also considering to incorporate explainability perspectives~\cite{ILRPA:23}, and validating it on a wider range of datasets and soil attributes, towards adoption in pedometrics and digital soil mapping applications.

\bibliographystyle{IEEEtran} 
\bibliography{references} 

\begin{thebibliography}{10}
\providecommand{\url}[1]{#1}
\csname url@samestyle\endcsname
\providecommand{\newblock}{\relax}
\providecommand{\bibinfo}[2]{#2}
\providecommand{\BIBentrySTDinterwordspacing}{\spaceskip=0pt\relax}
\providecommand{\BIBentryALTinterwordstretchfactor}{4}
\providecommand{\BIBentryALTinterwordspacing}{\spaceskip=\fontdimen2\font plus
\BIBentryALTinterwordstretchfactor\fontdimen3\font minus
  \fontdimen4\font\relax}
\providecommand{\BIBforeignlanguage}[2]{{%
\expandafter\ifx\csname l@#1\endcsname\relax
\typeout{** WARNING: IEEEtran.bst: No hyphenation pattern has been}%
\typeout{** loaded for the language `#1'. Using the pattern for}%
\typeout{** the default language instead.}%
\else
\language=\csname l@#1\endcsname
\fi
#2}}
\providecommand{\BIBdecl}{\relax}
\BIBdecl

\bibitem{keesstra2016forum}
S.~D. Keesstra, J.~Bouma, J.~Wallinga, P.~Tittonell, P.~Smith, A.~Cerd{\`a},
  L.~Montanarella, J.~Quinton, Y.~Pachepsky, W.~H. Van Der~Putten
  \emph{et~al.}, ``Forum paper: The significance of soils and soil science
  towards realization of the un sustainable development goals (sdgs),''
  \emph{Soil Discussions}, vol. 2016, pp. 1--28, 2016.

\bibitem{mcbratney1986introduction}
A.~McBratney, ``Introduction to pedometrics: a course of lectures,''
  \emph{CSIRO Australia, Division of Soils Technical Memorandum}, vol.~53, p.
  1986, 1986.

\bibitem{mcbratney2003digital}
A.~B. McBratney, M.~M. Santos, and B.~Minasny, ``On digital soil mapping,''
  \emph{Geoderma}, vol. 117, no. 1-2, pp. 3--52, 2003.

\bibitem{wadoux2020machine}
A.~M.-C. Wadoux, B.~Minasny, and A.~B. McBratney, ``Machine learning for
  digital soil mapping: Applications, challenges and suggested solutions,''
  \emph{Earth-Science Reviews}, vol. 210, p. 103359, 2020.

\bibitem{pennock2007soil}
D.~Pennock, T.~Yates, and J.~Braidek, ``Soil sampling designs,'' \emph{Soil
  sampling and methods of analysis}, pp. 1--14, 2007.

\bibitem{heuvelink2018uncertainty}
G.~B. Heuvelink, ``Uncertainty and uncertainty propagation in soil mapping and
  modelling,'' \emph{Pedometrics}, pp. 439--461, 2018.

\bibitem{torgo1996regression}
L.~Torgo and J.~Gama, ``Regression by classification,'' in \emph{Advances in
  Artificial Intelligence: 13th Brazilian Symposium on Artificial Intelligence,
  SBIA'96 Curitiba, Brazil, October 23--25, 1996 Proceedings 13}.\hskip 1em
  plus 0.5em minus 0.4em\relax Springer, 1996, pp. 51--60.

\bibitem{weiss1995rule}
S.~M. Weiss and N.~Indurkhya, ``Rule-based machine learning methods for
  functional prediction,'' \emph{Journal of Artificial Intelligence Research},
  vol.~3, pp. 383--403, 1995.

\bibitem{zhang2023improving}
S.~Zhang, L.~Yang, M.~B. Mi, X.~Zheng, and A.~Yao, ``Improving deep regression
  with ordinal entropy,'' in \emph{The Eleventh International Conference on
  Learning Representations}, 2023.

\bibitem{stewart2023regression}
L.~Stewart, F.~Bach, Q.~Berthet, and J.-P. Vert, ``Regression as
  classification: Influence of task formulation on neural network features,''
  in \emph{International Conference on Artificial Intelligence and
  Statistics}.\hskip 1em plus 0.5em minus 0.4em\relax PMLR, 2023, pp.
  11\,563--11\,582.

\bibitem{pintea2023step}
S.~L. Pintea, Y.~Lin, J.~Dijkstra, and J.~C. van Gemert, ``A step towards
  understanding why classification helps regression,'' in \emph{Proceedings of
  the IEEE/CVF International Conference on Computer Vision}, 2023, pp.
  19\,972--19\,981.

\bibitem{farebrother2024stop}
J.~Farebrother, J.~Orbay, Q.~Vuong, A.~A. Taiga, Y.~Chebotar, T.~Xiao,
  A.~Irpan, S.~Levine, P.~S. Castro, A.~Faust \emph{et~al.}, ``Stop regressing:
  Training value functions via classification for scalable deep rl,'' in
  \emph{Forty-first International Conference on Machine Learning}, 2024.

\bibitem{rothe2018deep}
R.~Rothe, R.~Timofte, and L.~Van~Gool, ``Deep expectation of real and apparent
  age from a single image without facial landmarks,'' \emph{International
  Journal of Computer Vision}, vol. 126, no.~2, pp. 144--157, 2018.

\bibitem{rogez2019lcr}
G.~Rogez, P.~Weinzaepfel, and C.~Schmid, ``Lcr-net++: Multi-person 2d and 3d
  pose detection in natural images,'' \emph{IEEE transactions on pattern
  analysis and machine intelligence}, vol.~42, no.~5, pp. 1146--1161, 2019.

\bibitem{zhang2016colorful}
R.~Zhang, P.~Isola, and A.~A. Efros, ``Colorful image colorization,'' in
  \emph{Computer Vision--ECCV 2016: 14th European Conference, Amsterdam, The
  Netherlands, October 11-14, 2016, Proceedings, Part III 14}.\hskip 1em plus
  0.5em minus 0.4em\relax Springer, 2016, pp. 649--666.

\bibitem{akkaya2019solving}
I.~Akkaya, M.~Andrychowicz, M.~Chociej, M.~Litwin, B.~McGrew, A.~Petron,
  A.~Paino, M.~Plappert, G.~Powell, R.~Ribas \emph{et~al.}, ``Solving rubik's
  cube with a robot hand,'' \emph{arXiv preprint arXiv:1910.07113}, 2019.

\bibitem{schrittwieser2020mastering}
J.~Schrittwieser, I.~Antonoglou, T.~Hubert, K.~Simonyan, L.~Sifre, S.~Schmitt,
  A.~Guez, E.~Lockhart, D.~Hassabis, T.~Graepel \emph{et~al.}, ``Mastering
  atari, go, chess and shogi by planning with a learned model,'' \emph{Nature},
  vol. 588, no. 7839, pp. 604--609, 2020.

\bibitem{quinlan1996bagging}
J.~R. Quinlan \emph{et~al.}, ``Bagging, boosting, and c4. 5,'' in
  \emph{Aaai/Iaai, vol. 1}, 1996, pp. 725--730.

\bibitem{clark1987induction}
P.~Clark, T.~Niblett \emph{et~al.}, ``Induction in noisy domains.'' in
  \emph{EWSL}, 1987, pp. 11--30.

\bibitem{hollmann2022tabpfn}
N.~Hollmann, S.~M{\"u}ller, K.~Eggensperger, and F.~Hutter, ``Tabpfn: A
  transformer that solves small tabular classification problems in a second,''
  in \emph{NeurIPS 2022 First Table Representation Workshop}, 2022.

\bibitem{mcelfresh2024neural}
D.~McElfresh, S.~Khandagale, J.~Valverde, V.~Prasad~C, G.~Ramakrishnan,
  M.~Goldblum, and C.~White, ``When do neural nets outperform boosted trees on
  tabular data?'' \emph{Advances in Neural Information Processing Systems},
  vol.~36, 2024.

\bibitem{muller2021transformers}
S.~M{\"u}ller, N.~Hollmann, S.~P. Arango, J.~Grabocka, and F.~Hutter,
  ``Transformers can do bayesian inference,'' in \emph{International Conference
  on Learning Representations}, 2022.

\bibitem{grinsztajn2022tree}
L.~Grinsztajn, E.~Oyallon, and G.~Varoquaux, ``Why do tree-based models still
  outperform deep learning on typical tabular data?'' \emph{Advances in neural
  information processing systems}, vol.~35, pp. 507--520, 2022.

\bibitem{li2021deep}
R.~Li, B.~J. Reich, and H.~D. Bondell, ``Deep distribution regression,''
  \emph{Computational Statistics \& Data Analysis}, vol. 159, p. 107203, 2021.

\bibitem{gawlikowski2023survey}
J.~Gawlikowski, C.~R.~N. Tassi, M.~Ali, J.~Lee, M.~Humt, J.~Feng, A.~Kruspe,
  R.~Triebel, P.~Jung, R.~Roscher \emph{et~al.}, ``A survey of uncertainty in
  deep neural networks,'' \emph{Artificial Intelligence Review}, vol.~56, no.
  Suppl 1, pp. 1513--1589, 2023.

\bibitem{hullermeier2021aleatoric}
E.~H{\"u}llermeier and W.~Waegeman, ``Aleatoric and epistemic uncertainty in
  machine learning: An introduction to concepts and methods,'' \emph{Machine
  learning}, vol. 110, no.~3, pp. 457--506, 2021.

\bibitem{schmidinger2023validation}
J.~Schmidinger and G.~B. Heuvelink, ``Validation of uncertainty predictions in
  digital soil mapping,'' \emph{Geoderma}, vol. 437, p. 116585, 2023.

\bibitem{kasraei2021quantile}
B.~Kasraei, B.~Heung, D.~D. Saurette, M.~G. Schmidt, C.~E. Bulmer, and
  W.~Bethel, ``Quantile regression as a generic approach for estimating
  uncertainty of digital soil maps produced from machine-learning,''
  \emph{Environmental Modelling \& Software}, vol. 144, p. 105139, 2021.

\bibitem{lagacherie2020analysing}
P.~Lagacherie, D.~Arrouays, H.~Bourennane, C.~Gomez, and L.~Nkuba-Kasanda,
  ``Analysing the impact of soil spatial sampling on the performances of
  digital soil mapping models and their evaluation: A numerical experiment on
  quantile random forest using clay contents obtained from vis-nir-swir
  hyperspectral imagery,'' \emph{Geoderma}, vol. 375, p. 114503, 2020.

\bibitem{sethi2017reliable}
T.~S. Sethi and M.~Kantardzic, ``On the reliable detection of concept drift
  from streaming unlabeled data,'' \emph{Expert Systems with Applications},
  vol.~82, pp. 77--99, 2017.

\bibitem{kakhani2024uncertainty}
N.~Kakhani, S.~Alamdar, N.~M. Kebonye, M.~Amani, and T.~Scholten, ``Uncertainty
  quantification of soil organic carbon estimation from remote sensing data
  with conformal prediction,'' \emph{Remote Sensing}, vol.~16, no.~3, p. 438,
  2024.

\bibitem{vaysse2017using}
K.~Vaysse and P.~Lagacherie, ``Using quantile regression forest to estimate
  uncertainty of digital soil mapping products,'' \emph{Geoderma}, vol. 291,
  pp. 55--64, 2017.

\bibitem{vogel2022direct}
S.~Vogel, E.~B{\"o}necke, C.~Kling, E.~Kramer, K.~L{\"u}ck, G.~Philipp,
  J.~R{\"u}hlmann, I.~Schr{\"o}ter, and R.~Gebbers, ``Direct prediction of
  site-specific lime requirement of arable fields using the base neutralizing
  capacity and a multi-sensor platform for on-the-go soil mapping,''
  \emph{Precision Agriculture}, vol.~23, pp. 127--149, 2022.

\bibitem{tavakoli2022rapidmapper}
H.~Tavakoli, J.~Correa, S.~Vogel, and R.~Gebbers, ``Rapidmapper--a mobile
  multi-sensor platform for the assessment of soil fertility in precision
  agriculture,'' \emph{AgEng LAND. TECHNIK}, pp. 351--58, 2022.

\bibitem{hengl2004generic}
T.~Hengl, G.~B. Heuvelink, and A.~Stein, ``A generic framework for spatial
  prediction of soil variables based on regression-kriging,'' \emph{Geoderma},
  vol. 120, no. 1-2, pp. 75--93, 2004.

\bibitem{gneiting2007strictly}
T.~Gneiting and A.~E. Raftery, ``Strictly proper scoring rules, prediction, and
  estimation,'' \emph{Journal of the American statistical Association}, vol.
  102, no. 477, pp. 359--378, 2007.

\bibitem{breure2022loss}
T.~S. Breure, S.~Haefele, J.~A. Hannam, R.~Corstanje, R.~Webster,
  S.~Moreno-Rojas, and A.~Milne, ``A loss function to evaluate agricultural
  decision-making under uncertainty: a case study of soil spectroscopy,''
  \emph{Precision Agriculture}, vol.~23, no.~4, pp. 1333--1353, 2022.

\bibitem{ILRPA:23}
M.~Iferroudjene, C.~Lonjarret, C.~Robardet, M.~Plantevit, and M.~Atzmueller,
  ``{Methods for Explaining Top-N Recommendations Through Subgroup
  Discovery},'' \emph{Data Mining and Knowledge Discovery}, vol.~37, pp.
  833--872, 2023.

\end{thebibliography}

\end{document}